\documentclass[9pt,conference]{IEEEtran}
\IEEEoverridecommandlockouts
\usepackage{cite}
\usepackage{amsmath,amssymb,amsfonts}
\usepackage[ruled]{algorithm2e}
\usepackage[dvipsnames]{xcolor}
\usepackage{graphicx}
\usepackage{pgfplots, tikz}
\usepackage{caption}
\usepackage{subcaption}
\usepackage{textcomp}
\usepackage{threeparttable}
\usepackage{multirow}
\usepackage{pifont}
\usepackage{listings}

\def\BibTeX{{\rm B\kern-.05em{\sc i\kern-.025em b}\kern-.08em
    T\kern-.1667em\lower.7ex\hbox{E}\kern-.125emX}}

\newcommand{\cmark}{\ding{51}}%
\newcommand{\xmark}{\ding{55}}%

\pgfplotsset{compat=1.18}

\begin{document}

\RestyleAlgo{ruled}

\title{MiCo: End-to-End Mixed Precision Neural Network Co-Exploration Framework for Edge AI}

\author{
\IEEEauthorblockN{Zijun Jiang and Yangdi Lyu$^{\dagger}$}
\IEEEauthorblockA{Microelectronics Thrust, The Hong Kong University of Science and Technology (Guangzhou)\\
$^{\dagger}$Corresponding author: yangdilyu@hkust-gz.edu.cn}
}

\maketitle

\begin{abstract}
Quantized Neural Networks~(QNN) with extremely low-bitwidth data have proven promising in efficient storage and computation on edge devices. To further reduce the accuracy drop while increasing speedup, layer-wise mixed-precision quantization~(MPQ) becomes a popular solution. However, existing algorithms for exploring MPQ schemes are limited in flexibility and efficiency. Comprehending the complex impacts of different MPQ schemes on post-training quantization and quantization-aware training results is a challenge for conventional methods. Furthermore, an end-to-end framework for the optimization and deployment of MPQ models is missing in existing work.

In this paper, we propose the MiCo framework, a holistic MPQ exploration and deployment framework for edge AI applications. The framework adopts a novel optimization algorithm to search for optimal quantization schemes with the highest accuracies while meeting latency constraints. Hardware-aware latency models are built for different hardware targets to enable fast explorations. After the exploration, the framework enables direct deployment from PyTorch MPQ models to bare-metal C codes, leading to end-to-end speedup with minimal accuracy drops.

\end{abstract}


\section{Introduction}

Tiny machine learning (ML) and edge artificial intelligence (AI) are becoming increasingly important and valuable in today's AI ecosystem. However, deploying AI models on edge devices is challenging due to the tight resource constraints. To address this, the quantization technique that converts high-precision floating-point data to low-precision integer data is widely adopted to reduce memory costs and improve the performance of edge AI models. For many AI applications, the quantization of neural networks can even reduce the data bitwidth to an extremely low range of below 4 bits, while still maintaining acceptable accuracy.

Instead of using a uniform bitwidth for all layers, layer-wise mixed precision quantization~(MPQ) assigns different precisions to each model layer to achieve better accuracy and efficiency. To retain the accuracy while accelerating the inference of the quantized model, it is essential to design MPQ schemes by considering layer sensitivity and computational cost. As the number of layers keeps increasing in recent AI models, manually analyzing and assigning MPQ schemes is becoming impractical. Search algorithms with heuristic strategies~\cite{hawqV3,ZeroQ} and machine learning approaches~\cite{haq,SDQ,van2023bomp} have been developed to explore MPQ schemes and improve MPQ model performance, but the high dimensionality and complex inter- and intra-layer correlation of the MPQ search spaces make the algorithms less efficient. Furthermore, as the bitwidths become smaller, performing quantization-aware training~(QAT) becomes necessary, while its time-consuming nature restricts the number of schemes to evaluate within a limited time budget.

\begin{table}[ht]
    \centering
    \begin{threeparttable}
    \caption{State-of-the-art Mixed Precision Hardware}
    \label{tab:MP_Hardware}
    \begin{tabular}{ c | c | l | l }
    \hline
    \textbf{Type} & \textbf{Hardware} & \textbf{Bitwidths} & \textbf{Perf.(GOPs)} \\
    \hline
    \multirow{2}{*}{\textbf{Accelerator}} & BitFusion~\cite{bit_fusion} & 8b-2b &  - \\
    & SPEED~\cite{speed} & 16b/8b/4b & 343.1-737.9 \\
    \hline
    \multirow{5}{*}{\textbf{CPU Extension}} & XpulpNN~\cite{xpulpnn} & 16/8b/4b/2b   & 19.0-48.0 \\
    & MPIC~\cite{MPIC}             & 8b/4b/2b      & 1.1-3.3 \\
    & Mixed-GEMM~\cite{mix-gemm}   & 8b-2b         & 4.2-7.9   \\
    & MPQ VMM~\cite{edge-mpq}      & 16b/8b/4b/2b  & 0.93-2.86 \\
    & Multi-Pump~\cite{multi-pump} & 8b/4b/2b      & 0.24-0.85 \\
    \hline
    \end{tabular}
    \end{threeparttable}
\end{table}

Besides, the efficient deployment of MPQ models on specific hardware platforms becomes another challenge. As shown in Tab.~\ref{tab:MP_Hardware}, the existing mixed-precision hardware is diverse in architectures and bitwidth supports. Some of them are application-specific accelerators, such as~\cite{bit_fusion, speed,unpu}, which are designed to support mixed-precision computation with high performance. However, they are usually loosely coupled to processors and occupy a large silicon area, which is not desirable for edge devices. A more appropriate solution for edge environments is to extend CPUs with custom instruction sets and tightly integrated units, given their lower area and energy overhead. Existing works like~\cite{MPIC, edge-mpq} extend RISC-V CPUs in a single-instruction multiple-data~(SIMD) manner, which enables parallel computation of multiple low-bitwidth data packed in a 32- or 64-bit word. 

While these hardware platforms are capable of accelerating mixed-precision computation, existing end-to-end deployment frameworks, such as TVM~\cite{TVM} and DORY~\cite{dory}, primarily focus on INT8 models and do not adequately support lower bitwidths or mixed precision models. A holistic hardware-aware flow that efficiently searches the MPQ schemes and deploys the MPQ models on various hardware targets is still absent. 

To address these challenges in the MPQ field, in this work, we aim to develop a framework to explore optimal MPQ schemes for given AI models with a better awareness of the underlying hardware, within the limitation of search budgets.

The main contributions of this work are as follows:
\begin{itemize}
    \item A novel exploration algorithm for searching MPQ schemes for neural networks under certain constraints. By combining ensemble learning models (random forests) and specially designed sampling strategies, our approach enables efficient exploration of MPQ schemes, leading to lower accuracy drops under the same constraints. 
    \item A hardware-aware latency proxy modelling method that can adapt to different hardware targets. The proposed proxy model produces latency estimates that more accurately reflect the performance of real hardware compared to the traditional BOPs metric, effectively guiding the MPQ exploration towards the schemes that deliver true acceleration.
    \item A holistic framework to explore MPQ models and deploy them onto various hardware platforms, including accelerators and CPUs with mixed-precision acceleration support. We illustrate this framework through two case studies on different hardware platforms, where speedup is achieved by the MPQ models with minimal accuracy drops.
\end{itemize}

\section{Background}

\subsection{Mixed Precision Quantization}
\label{sec:MPQ}

The most common quantization approach is to quantize the entire network into a fixed precision, e.g., INT8 or INT4. However, applying a uniform precision across all layers can create a trade-off: a high compression ratio often leads to significant accuracy degradation, while a low compression ratio results in limited performance gains. The reason behind this is that layers in neural networks have various impacts on the final output; some of them are sensitive to quantization errors and require higher precisions, while others are robust even with extremely low precisions. 

Therefore, layer-wise mixed precision quantization that allows different precisions for layers is introduced to add flexibility and improve the efficiency of the QNNs. In each layer, the quantization can be applied to both weights and activations, converting high-precision floating-point data to low-precision integer data. The quantization configuration of each layer can be noted as W$x$A$y$, where $x$ and $y$ represent the bit widths of weights~($b_w$) and activations~($b_a$), respectively. For example, a layer with its weights quantized to 4-bit and its activations quantized to 8-bit can be denoted as a layer with an MPQ scheme of W4A8.

As the number of layers in a network increases, the search space for MPQ schemes grows exponentially. With $B$ possible schemes for each layer, the total space can be $B^L$ networks for a model with $L$ layers. This exponential growth motivates innovation in efficient searching methods.

\subsection{PTQ v.s. QAT}
\label{sec:ptq_vs_qat}
Post-training quantization~(PTQ) and quantization-aware training~(QAT) are two main approaches to creating an MPQ model. With a model pre-trained with high-precision variables, PTQ directly quantizes the model using the chosen MPQ schemes, while QAT runs a re-training process on the MPQ model. PTQ is straightforward and time-efficient, but often leads to lower accuracy in the resulting models. In contrast, QAT demands more computational resources and a longer training period, but it typically yields models with improved accuracy.

When compression is pushed to extremely low bitwidths of less than 4 bits, the accuracy drop for PTQ becomes substantial, making QAT essential for achieving acceptable accuracy levels. In Fig.~\ref{fig:ptq_vs_qat}, we demonstrate the difference between QAT accuracy and PTQ accuracy for a convolutional neural network (CNN) model trained on the CIFAR-10 dataset, when one of its layers is quantized to different weight and activation precisions.
\definecolor{QAT}{HTML}{94FFD8}
\definecolor{PTQ}{HTML}{A3D8FF}

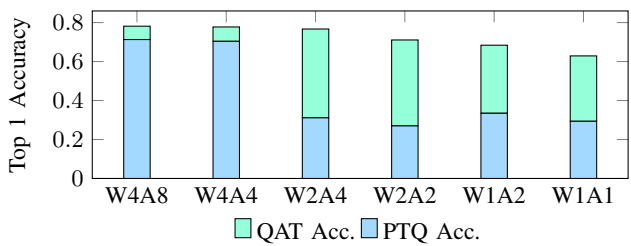
\begin{figure}[h]
\begin{tikzpicture}
\begin{axis}[
    ybar stacked,    
    height=1.5in,
    width = 0.48\textwidth,
    ylabel={Top 1 Accuracy},
    ymin=0,
    xtick={1,2,3,4,5,6},
    xticklabels={%
        W4A8,
        W4A4,
        W2A4,
        W2A2,
        W1A2,
        W1A1},
    reverse legend,
    legend style={at={(0.5,-0.2)},
    anchor=north,legend columns=-1, draw=none},
    y label style={at={(-0.1,.5)},anchor=south}]

    \addplot [fill=PTQ] coordinates {
        (1, 0.7129) (2,0.7039) (3, 0.3111) (4, 0.2697) (5, 0.3347) (6, 0.2935)
    };

    \addplot [fill=QAT] coordinates {
        (1, 0.0686) (2,0.0737) (3, 0.4559) (4, 0.4411) (5, 0.3492) (6, 0.3355)
    };
    
    \legend{PTQ Acc., QAT Acc.}
    
\end{axis}
\end{tikzpicture}
\caption{QAT Accuracy v.s. PTQ Accuracy}
\label{fig:ptq_vs_qat}
\end{figure}
It can be observed that the accuracy difference between QAT and PTQ is small for bitwidths above 4~(e.g., W4A8 and W4A4), but becomes significant for bitwidths below 4. Therefore, when exploring the MPQ search space, it is necessary to handle these two bitwidth cases~(above 4-bit and below 4-bit) with different approaches.

\subsection{Hardware-aware MPQ Exploration}
\label{sec:mpq_bops}
To efficiently explore the MPQ schemes with high accuracy and performance, it is essential to estimate the inference latencies of MPQ models on specific hardware targets. However, the latency results are typically obtained from cycle-accurate simulators, which take seconds to minutes to run, depending on the hardware architectures and the complexity of the model. Since hardware latencies are frequently evaluated during the exploration, a metric called Bit Operations (BOPs) was introduced in previous works~\cite{uniq} to serve as an indicator for the computational cost of mixed precision models, thereby avoiding the time-consuming simulations. 

The BOPS for a layer in the model are defined as follows:
\begin{equation}
    \text{BOPs}_i = b_{w_i} b_{a_i} \text{MACs}_i
    \label{eq:bops}
\end{equation}
where $b_{w_i}$ and $b_{a_i}$ represent the bitwidths for weights and activations, respectively, and $\text{MACs}_i$ is the number of MAC operations of the $i$-th layer.

Many existing MPQ frameworks~\cite{hawqV3, bayesian_bits, edge-mpq} use BOPs as a proxy for hardware latency, facilitating efficient hardware-aware exploration. The MPQ exploration can be formulated as an optimization problem as follows:
\begin{equation}
    \begin{aligned}
    \text{Maximize: }   & \text{Accuracy}(M_Q) \\
    \text{Subject to: } & \sum_{i=1}^L \text{BOPs}_i \leq \text{BOPs}_{\text{constr}}\text{ (Bit Operation Constraint)} \\
    \end{aligned}
\label{eq:opt}
\end{equation}
where $M_Q$ is a MPQ quantized version of the neural network model $M$ with a quantization scheme $Q$. This scheme $Q$ is represented as $[(b_{w_1}, b_{a_1}), \cdots, (b_{w_L}, b_{a_L})]$, which consists of a vector of bit width pairs for weights and activations, with $L$ denoting the number of layers in $M$. The objective is to find a quantization scheme $Q$ that maximizes the accuracy of $M_Q$ while ensuring the BOPs meet the specified constraints. However, this optimization problem is challenging to solve due to the exponential growth of the search space for $Q$, which increases as $B^L$ with the number of layers $L$ and the available configurations $B$ for each layer. Consequently, existing approaches typically employ heuristic algorithms or machine learning techniques to explore the MPQ spaces, allowing the search algorithms to complete within the limitations of time and computational resources.

Despite the wide use of BOPs in MPQ exploration, it has several limitations when it comes to end-to-end optimization. First, while BOPs serve as a metric for computational complexity, they do not consider other essential hardware-related factors that impact performance, such as memory usage. Second, the performance gain from the reduction of BOPs can vary significantly depending on hardware architectures. These limitations lead to unsatisfactory results in MPQ exploration, ultimately hindering the acceleration.

\begin{figure}[b]
    \centering
    \begin{subfigure}{0.24\textwidth}
    \includegraphics[width=\textwidth]{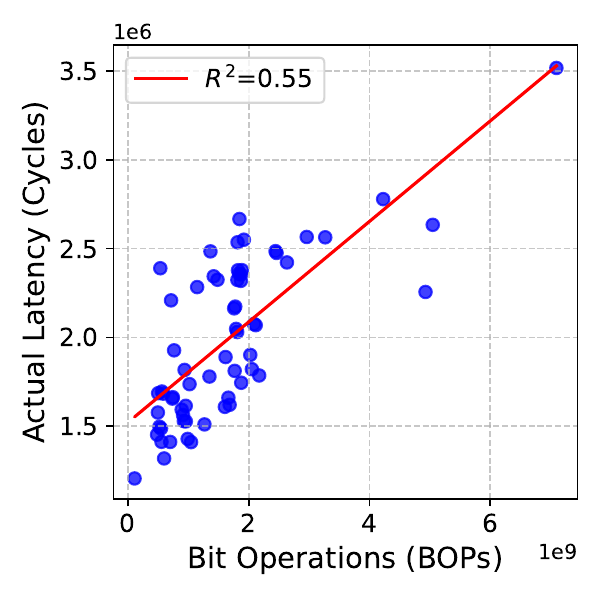}
    \caption{VGG7 on BitFusion}
    \label{fig:latency_on_accelerator}
    \end{subfigure}
    \begin{subfigure}{0.24\textwidth}
    \includegraphics[width=\textwidth]{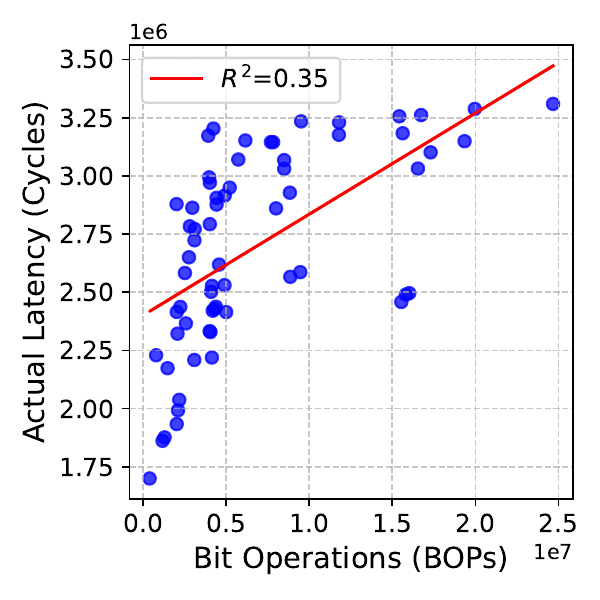}
    \caption{LeNet5 on Extended CPU}
    \label{fig:latency_on_cpu}
    \end{subfigure}
    \caption{Correlation Between BOPs and Actual Latency}
    \label{fig:corr_bops}
\end{figure}

To illustrate these limitations, we randomly sampled 64 MPQ schemes on the VGG7~\cite{vgg} and LeNet5~\cite{lenet} models, and measured their actual latencies on a mixed-precision accelerator (BitFusion) and a SIMD-extended CPU, respectively. The results are illustrated in Fig.~\ref{fig:corr_bops}, which shows the correlations between BOPs and the actual latency of MPQ models.

As shown in Fig.~\ref{fig:corr_bops}, although the BOPs metric shows positive correlation with the actual latency, the $R^2$s are relatively low, which indicates that relying solely on BOPs to optimize or predict latency may not be effective. Therefore, there is a need for improved hardware proxies to achieve speedup in end-to-end latency.

\section{Mixed Precision Quantization Exploration}
\label{sec:exploration flow}
We propose to solve the MPQ exploration problem using ensemble learning models that predict accuracy results from the MPQ schemes, and focusing on near-constraint spaces to improve the efficiency of the search for optimal MPQ schemes. Moreover, we implement a sampling method to generate initial training samples to augment the information learned by the accuracy models, which further enhances the MPQ search efficiency.

\begin{figure}[ht]
    \centering
    \includegraphics[width=0.48\textwidth]{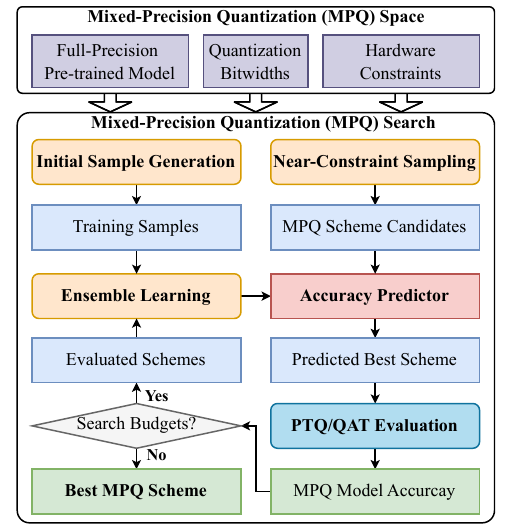}
    \caption{MiCo MPQ Search Flow}
    \label{fig:search flow}
\end{figure}

The basic flow of MPQ search in our framework is shown in Fig.~\ref{fig:search flow}. The process begins with a pre-trained neural network model, the available bitwidth options, and specified BOPs or latency constraints along with a search budget. The MPQ search starts with a particular set of initial samples. After the initial sampling, the algorithm develops an accuracy predictor that learns from the sampled data~(i.e., MPQ schemes and their corresponding accuracy results). It then refines the predictive model to estimate the accuracies of various schemes and recommends new MPQ schemes for evaluation. The process forms a classical optimization loop, which ends when running out of budget. Finally, the explored scheme with the highest accuracy from PTQ or QAT evaluation is selected for deployment.

\begin{table}[ht]
    \centering
    \caption{Comparison with Existing Algorithms}
    \label{tab:SW_Cmp}
    \begin{tabular}{ l | c l c c}
    \hline
    \textbf{Work}  & \textbf{Method} & \textbf{Bitwidths}& Support $b_a \neq b_w$? & $B$ \\
    \hline
    HAWQ-V3~\cite{hawqV3}     & ILP & 8/4        & \xmark                & 2    \\
    $w$-based~\cite{edge-mpq} & NLP & 8/7/6/5/4  & \xmark                & 5    \\
    BOMP~\cite{van2023bomp}   & BO  & 8/7/6/5/4  & \xmark                & 5    \\
    HAQ~\cite{haq}            & RL  & 8/6/4/2    & \color{teal}{\cmark}  & 16   \\
    \hline
    \textbf{MiCo}~(PTQ)       & RF  & 8/7/6/5/4   & \color{teal}{\cmark} & \textbf{25}     \\
    \textbf{MiCo}~(QAT)       & RF  & 8/4/2/1 & \color{teal}{\cmark} & \textbf{16}     \\
    \hline
    \end{tabular}
\end{table}

A comparison of the methods and supported scheme choices $B$ of existing frameworks is listed in Tab.~\ref{tab:SW_Cmp}. The first two approaches are based on Integer Linear Programming~(ILP) and Non-linear Programming~(NLP) methods, which heuristically solve the MPQ problem with layer-wise model statistics, such as layer-wise sensitivity and bit operations. However, these approaches are limited in their quantization choices as they assume uniform bit widths for weights and activations across each layer. BOMP~\cite{van2023bomp}, on the other hand, utilizes Gaussian processes~(GP) and Bayesian optimization to search for MPQ schemes. While it shows high searching efficiency compared to other methods, Gaussian process models become less efficient when dealing with high-dimensional problems, such as deep neural networks with a large number of layers. Finally, HAQ~\cite{haq} leverages Reinforcement Learning (RL) techniques, where agents observe and learn the optimal quantization choices for each layer. Although HAQ allows more choices by supporting mixed weight and activation bitwidths, it requires significantly more samples for the RL agents to establish the relationship between schemes and accuracy results, which can be time-consuming. 

In our approach, we employ ensemble random forest models to capture the complex correlation between MPQ schemes and their resulting PTQ/QAT accuracies. Additionally, we implement several specially designed sampling techniques to optimize the MPQ schemes while adhering to BOPs and latency constraints.

\subsection{Ensemble Learning-based MPQ Accuracy Predictor}

To evaluate the accuracy of an MPQ model with a given quantization scheme $Q$, it always requires validation on the testing dataset, which can be both computationally intensive and time-consuming. Therefore, in the MPQ search problem, it is common practice to train an accuracy predictor on a small set of MPQ schemes and their corresponding accuracies. The predictor serves as a cost-effective proxy model for approximating the actual accuracies.

In our approach, we employ Random Forest~(RF)~\cite{random_forest}, an ensemble learning method well-suited for regression tasks like accuracy prediction. RF operates by constructing a large number of decision trees during training and providing the mean prediction (regression) from these individual trees. Each tree is trained on a random subset of the data and considers only a random subset of features at each split. This strategy helps to reduce variance and mitigate overfitting compared to a single decision tree. 

RF is well-suited to the challenges of MPQ exploration: navigating a high-dimensional, discrete search space with complex non-linear accuracy relationships under tight PTQ/QAT evaluation budgets. Specifically, RF scales better computationally in high dimensions than alternatives like GPs, effectively models non-linear interactions between discrete bitwidth features, and provides robust predictions from relatively few samples. This sample efficiency is crucial given expensive accuracy evaluations and offers an advantage over RL approaches, which are typically more data-hungry. While GPs provide uncertainty estimates, their cost and kernel sensitivity can be limiting.

In recent advances in network architecture search~(NAS)~\cite{NAS_weak, gp_nas_ensemble}, it is proved that with well-designed learning strategies, the ensemble learning models like random forest can deliver good search results. Inspired by this philosophy, we not only build our accuracy predictors with random forests for the MPQ problem, but also try to boost their optimization ability with techniques specially designed for MPQ problems.

\subsection{Layer-wise Orthogonal Initial Sampling}
\label{sec:initial}
For learning-based algorithms, the representative initial samples can strengthen the performance of predictive models like random forest models and Gaussian process models. A common way of generating initial samples is through uniform random sampling of the search space. However, we have found that random sampling is very inefficient for MPQ problems due to their extremely huge space.

To address this issue, we propose a layer-wise orthogonal initial sampling method, which is described in Algorithm~\ref{alg:init}. First, to make sure the algorithm learns the upper and lower bounds of accuracy, we include the MPQ schemes with the highest and lowest bitwidths~(e.g., W8A8 and W1A1) in the initial samples. Then, for each subsequent initial sample, the method randomly selects several layers and quantizes them with random schemes, while keeping other layers in the W8A8 scheme. The random selection guarantees that every layer is quantized at least once, and that the selected layers in each initial sample do not overlap, forming a set of orthogonal samples.
\begin{algorithm}[ht]
\caption{Layer-wise Orthogonal Initial Sampling}\label{alg:init}
    \KwIn{$N_{init}$-Number of Initial Samples, $M$-MPQ Model}
    \KwOut{$\mathcal{S}$-Initial Samples}
    $\mathcal{S} \gets \emptyset$; \\
    $b_{max} \gets$ Largest supported bitwidth; \\
    $b_{min} \gets$ Lowest supported bitwidth; \\
    $Q_1 \gets [(b_{max},b_{max}),(b_{max},b_{max}),\cdots,(b_{max}, b_{max})]$ \\
    $Q_2 \gets [(b_{min},b_{min}),(b_{min},b_{min}),\cdots,(b_{min}, b_{min})]$ \\
    Randomly permute the layer indices and get $[l_1, l_2, \cdots, l_L]$\\
    $N_m \gets \max(L/(N_{init}-2),1)$; \\
    \For{$i=3 \to N_{init}$}{
        $Q_i \gets [(b_{max},b_{max}),(b_{max},b_{max}),\cdots,(b_{max}, b_{max})]$ \\
        Modify the layers starting from $start = (i - 3) * N_m$\\
        \For{$k=1 \to N_{m}$}{
            Randomly modify bitwidths of the layer $Q_i[l_{start + k}]$; \\
        }
    }
    \For{$i=1 \to N_{init}$}{
        $Acc_i \gets$ Evaluate $M_{Q_i}$ on the test dataset; \\
        $\mathcal{S} \cup (Q_i, Acc_i)$
    }
    \Return $\mathcal{S}$
\end{algorithm}

With the above sampling method, the effects of different quantized layers and quantization schemes on accuracy are effectively captured in the initial samples, enabling our prediction model to acquire valuable insights.

\subsection{Near-constraint Scheme Optimization}
\label{sec:optimization}



\begin{figure}[htbp]
    \centering
    \includegraphics[width=0.45\textwidth]{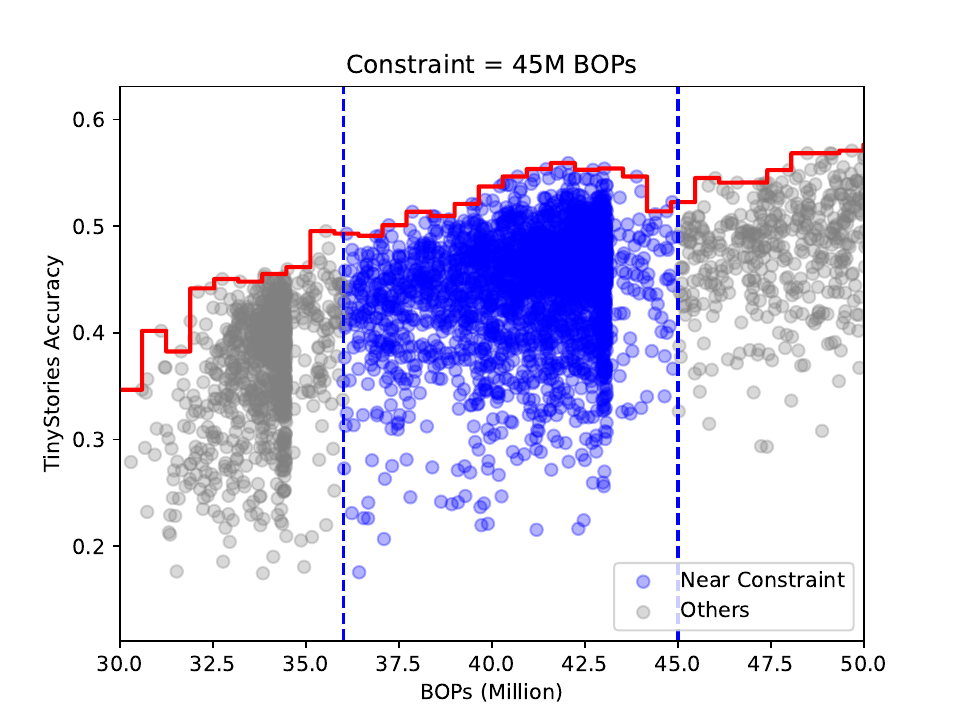}
    \caption{Accuracy v.s. BOPs for TinyLLaMa on TinyStories}
    \label{fig:acc_vs_bops}
\end{figure}

Consider the optimization problem with equality constraints version of Eq.~\ref{eq:opt}, defined as $F(x) = max(Acc(M_Q))$ subject to $\sum \text{BOPs} = x$. Our goal is to find the maximum value of $F(x)$ under the constraint that $x \leq \text{BOPs}_{\text{constr}}$. As a result, only the schemes that meet the constraint should be evaluated. We observe that $F(x)$ is positively correlated with the BOPs value $x$, although it exhibits slight fluctuations. This aligns with the intuitive understanding that higher precision and greater computational resources typically lead to improved accuracy. To validate it in a real design, we conducted an experiment by randomly sampling MPQ schemes on the TinyLLaMa model~(details in Tab.~\ref{tab:networks}), with the results illustrated in Fig.~\ref{fig:acc_vs_bops}. The test accuracy and the corresponding BOPs are represented by dots, while $F(x)$ is shown as a red line. As illustrated, $F(x)$ tends to increase with BOPs.

This motivates us to focus the search space near the constraint. We integrate a simple genetic algorithm to generate samples within a certain range, which evaluates the predicted accuracy within the near-constraint space. Initially, the range is set between $0.8*\text{BOPs}_{\text{constr}}$ and $\text{BOPs}_{\text{constr}}$, and it is gradually expanded to the range between $0.5*\text{BOPs}_{\text{constr}}$ and $\text{BOPs}_{\text{constr}}$ towards the end of the exploration. This strategy prioritizes schemes close to the constraints to quickly obtain good solutions, while allowing for the exploration of slightly more distant areas in case we discard promising options. Additionally, it enhances efficiency by disregarding sub-optimal regions. For example, if the BOPs constraint is set at 45M BOPs, the initial focused range is shown as the blue points in Fig.~\ref{fig:acc_vs_bops}, spanning from 36M to 45M BOPs.

Besides, we also apply the common heuristic of maintaining high precision for the first and last layers in our sampling method, which has been proved to be effective in previous works~\cite{haq, multi-pump}.

\subsection{Accuracy Evaluation}

In each search iteration, the predicted best schemes are evaluated to get the actual accuracy of the MPQ models. In our flow, we use different evaluation processes for PTQ and QAT search. While PTQ accuracy is directly validated, the QAT accuracy is validated after a very short re-training~($\approx$ pre-trained time/20). This short re-training costs less time than a full re-training, while offering the algorithm a good approximation of the QAT recoverable accuracy mentioned in Sec.~\ref{sec:ptq_vs_qat}.

\section{End-to-End Exploration \& Deployment}

The MiCo framework enables the end-to-end exploration and deployment of MPQ models for various hardware targets. To achieve this, we incorporate a hardware-aware proxy model into the exploration flow explained in Sec.~\ref{sec:exploration flow} to improve the estimation of the real end-to-end latency. 

\begin{figure}[h]
    \centering
    \includegraphics[width=0.48\textwidth]{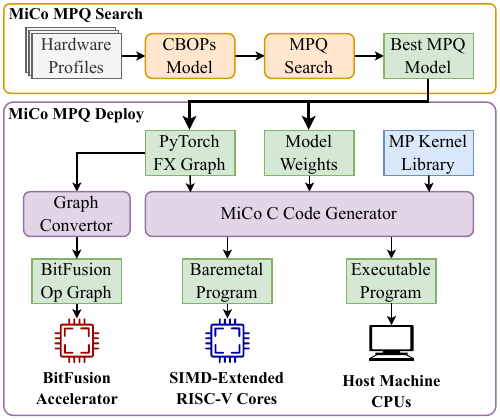}
    \caption{MiCo End-to-End Exploration \& Deployment}
    \label{fig:E2E_ED}
\end{figure}

The framework also offers complete deployment infrastructures, including MPQ model exporting, graph extraction, mixed-precision inference library, and code generation, which is shown in Fig.~\ref{fig:E2E_ED}. This creates a seamless transition from PyTorch models to executable bare-metal programs on edge devices.

\subsection{Hardware-aware Proxy Model}
\label{sec:cbops_proxy}


As discussed in Sec.~\ref{sec:mpq_bops}, while the BOPs metric is easy to estimate, its limitations make it less reliable for a truly hardware-aware exploration. To overcome the shortcomings of BOPs, we propose the \textbf{composite bit operations}~(CBOPs), which is a hardware-dependent proxy model modified from BOPs.

There are 3 main variables used to establish the CBOPs model: Bit Multiply-Accumulate Operations (BMACs), Activation Loads (ALoads), and Weight Loads (WLoads), which are defined as follows:
\begin{equation}
\begin{split}
    &\text{BMACs}_i = \max(b_{w_i}, b_{a_i}) \text{MACs}_i \\
    &\text{ALoads}_i = b_{a_i} \text{MACs}_i  \\
    &\text{WLoads}_i = b_{w_i} \text{MACs}_i
\end{split}
\end{equation}
BMACs represents the total number of bits involved in the underlying hardware MAC units, and ALoads and WLoads represent the number of bits for activations and weights loaded from the memory to perform the MAC operations. Notably, the BMACs is calculated by multiplying the maximum bitwidths of weights and activations by the number of MACs, since the parallelism of most mixed precision hardware is bounded by the larger operands. The other two variables, ALoads and WLoads, were selected to address the BOPs metric's inability to distinguish between the impacts of weight and activation bitwidths.

Using these 3 basic variables, hardware latency can be modelled with different techniques, which can be fitted to a small set of profiling data specific to the hardware target. For example, the CBOPs model may take the form of a linear regression model, which is defined as:
\begin{equation}
    \text{CBOPs}_i = \beta_{\text{M}}\text{BMACs}_i + \beta_{\text{A}}\text{ALoads}_i + \beta_{\text{W}}\text{WLoads}_i + C
    \label{eq:cbops}
\end{equation}
The hardware-dependent coefficients $\beta_{\text{M}}$, $\beta_{\text{A}}$, $\beta_{\text{W}}$, and the constant $C$ are determined using the benchmarking data to build the CBOPs model in Eq.~\ref{eq:cbops}.

For hardware where latency scales linearly (e.g., systolic arrays), linear CBOPs models (Eq.~\ref{eq:cbops}) tend to provide accurate predictions. However, in hardware platforms where latency exhibits non-linear behaviour (e.g., due to caching effects or specific operations like im2col on CPUs), other regressors like random forests or regression trees are more suitable to model the relationship between these variables and latency. Moreover, other factors like input shapes and weight shapes can also be used in the latency modelling to improve the latency prediction, but in this work, we focus solely on the three variables mentioned, as they represent the minimal set necessary for reliable predictions based on our observations.

For each hardware target with different architectures and mixed-precision performance, we begin the MPQ exploration by executing a small set of kernel benchmarks across every bitwidth to create the \textbf{hardware profiles}. After fitting the CBOPs models with this profiling data, these models are used in the following MPQ exploration as the proxy for hardware latency. Different CBOPs models are built for each computation kernel~(MatMul, Conv2D). Training the CBOPs models is a one-time effort for each hardware target, as the proxies can be used for the MPQ exploration of all the networks.

With CBOPs models, the hardware latency can be better reflected during the MPQ exploration, which guides the exploration towards the MPQ schemes with real accelerations.

\subsection{Multi-platform Deployment}

The lower part of Fig.~\ref{fig:E2E_ED} shows the deployment flow of the MiCo framework. The best MPQ model explored by the search flow is first parsed into a PyTorch FX graph, which represents how the model calls neural network functions, operators, and modules during inference. Then, depending on the specific hardware platform to deploy, the graph is converted into different forms that incorporate the MPQ scheme of the model.

Currently, our deployment flow supports 3 hardware platforms: the BitFusion accelerator~\cite{bit_fusion}, the custom SIMD-extended RISC-V CPU, and the host machine that runs the MiCo framework. The core infrastructures of our deployment flow are as follows:

\paragraph{Graph Convertor}
The simulator of BitFusion is based on DnnWeaver~\cite{dnnweaver}. Therefore, to simulate the MPQ model on BitFusion, we have developed a \textbf{graph convertor} that converts the PyTorch FX graph into a DnnWeaver operator graph, and assigns the MPQ scheme to each operation in the graph. This operator graph with bitwidth assignments is deployed and simulated on the BitFusion simulator.

\paragraph{Mixed Precision Computation Library}
For edge devices with CPUs, a bare-metal C code library is essential for the inference of edge AI models. While there are many C libraries like TVM~\cite{TVM} and CMSIS-NN~\cite{cmsis-cnn} that support floating-point and 8-bit integer data, there is limited support for sub-byte data (4-, 2-, and 1-bit). Therefore, we extend the library of baremetal-NN~\cite{baremetal-nn} to support the MPQ kernels (MatMul, Conv2D) with all mixed bitwidth combinations of 8, 4, 2, and 1-bit. Additionally, we have implemented the SIMD-accelerated variants of these MPQ kernels with our custom instructions~(details in Sec.~\ref{sec:e2e_on_rv}). To support the dynamic quantization of activations, we have also implemented miscellaneous functions like quantization, de-quantization, packing, and unpacking. With our MP kernel library, the complete inference of MPQ models can be fully supported.

\paragraph{MiCo C Code Generator}
For the deployment on CPUs, the PyTorch FX graph derived from the MPQ model is translated into C code by our C code generator. This generator traces all functions and operations involved in the forwarding pass of the model, and maps them into the kernels in the aforementioned MP library. Then, it allocates all the buffers for layer inputs and outputs. Finally, it exports the quantized model in binary format, and assigns the addresses to the weight pointers of each layer. The generated C code can be compiled and run on both RISC-V CPUs and the host machine CPUs (e.g., x86).

Additionally, the holistic end-to-end flow is wrapped into Python APIs, which allows us to run the exploration and deployment with only a few lines of Python scripts.


\section{Experimental Results}

\subsection{MPQ Search Setup}

Tab.~\ref{tab:networks} lists the various baseline models used in our experiments, sorted by their number of layers. 
CNN4 and LeNet5 have fewer than 5 layers and are trained on the CIFAR-10 and MNIST datasets, respectively. SqueezeNet consists of 26 layers and is trained on the CIFAR-100 dataset. Additionally, due to the great attention paid to generative language models in recent years, we also include a small-scale Transformer-based language model, TinyLLaMa, which has 1 million parameters and is trained on the TinyStories dataset~\cite{tinystories} for basic language generation tasks on edge. The training and validation of models are conducted on a server with an Nvidia A30 GPU, and all of the models and quantization are implemented with PyTorch.

\begin{table}[ht]
\centering
\begin{threeparttable}
    \caption{Baseline Full Precision Models}
    \begin{tabular}{| c | c | c | c |}
    \hline
    \textbf{Model} & \textbf{Acc.~(\%)} & \textbf{\# Layers}     & \textbf{MACS}\\
    \hline
    CNN4~\cite{cmsis-cnn}          & 75.79  & 3$\times$Conv2D, 1$\times$Linear & 12.3M        \\
    \hline
    LeNet5~\cite{lenet}            & 99.13  & 2$\times$Conv2D, 3$\times$Linear & 0.39M         \\ 
    \hline
    SqueezeNet~\cite{squeezenet}   & 63.58  & 26$\times$Conv2D  & 53.3M        \\
    \hline
    TinyLLaMa~\cite{llama2}        & 60.91\tnote{*}  & 36$\times$Linear  & 1.35M        \\
    \hline
    \end{tabular}
    \label{tab:networks}
\begin{tablenotes}
\item [*] For TinyLLaMa, the accuracy refers to next token prediction accuracy. 
\end{tablenotes}
\end{threeparttable}
\end{table}

We primarily compared our method with three state-of-the-art methods mentioned in Tab.~\ref{tab:SW_Cmp}, which include the non-linear programming-based $w$ method~\cite{edge-mpq}, the reinforcement learning-based HAQ method~\cite{haq}, and the Bayesian optimization-based BOMP~\cite{van2023bomp}. To facilitate this comparison, we directly integrated the open-source implementation of HAQ into our framework and re-implemented the other two methods based on their published descriptions.

For our method and two machine learning-based methods~(HAQ and BOMP), the sample budgets are set to the same numbers for fair comparisons. In the case of the $w$-based method, which performs static analysis of layer-wise sensitivities, the required number of samples equals the number of layers in each model.

We conduct the exploration under different BOPs constraints, which are defined as specific ratios of the base BOPs derived from INT8 models, where all $b_w$ and $b_a$ values are fixed at 8. We repeat each experiment with different seeds for each algorithm, and the average results are reported. 

\subsection{Mixed Precision PTQ Search Results}

As mentioned in Sec.~\ref{sec:ptq_vs_qat}, for higher bitwidths, PTQ will not lead to huge accuracy drops. Therefore, we first conducted the MPQ search with PTQ. The results of the PTQ search are shown in Tab.~\ref{tab:ptq search}, where the bitwidth choices are limited from 4 bits to 8 bits, which is the space originally supported in the $w$-based method~\cite{edge-mpq} and BOMP~\cite{van2023bomp}, as mentioned in Tab.~\ref{tab:SW_Cmp}. The search budgets for learning-based methods are set to 48 samples, consisting of 16 initial samples and 32 search samples.

\begin{table}[ht]
\centering
\caption{PTQ Search (4/5/6/7/8-bit) Accuracy Results}
\label{tab:ptq search}
\begin{tabular}{ l | l | c | c | c }
\hline
\textbf{Model}              & \textbf{Method} & \textbf{0.6BOPs} & \textbf{0.5BOPs} & \textbf{0.4BOPs} \\
\hline
\multirow{4}{*}{CNN4}       & $w$-based       & 71.58 & 71.68  & 71.23 \\
                            & HAQ             & 75.02 & 74.61  & 73.68 \\
                            & BO              & \textbf{75.29} & 75.12  & 74.86 \\
                            & \textbf{Ours}   & 75.26 & \textbf{75.19}  & \textbf{75.12} \\
\hline
\multirow{4}{*}{LeNet5}     & $w$-based       & 98.92 & 99.12 & 99.06 \\
                            & HAQ             & 99.12 & 99.10 & 99.09 \\
                            & BO              & 99.16 & 99.15 & 99.16 \\
                            & \textbf{Ours}   & \textbf{99.17} & \textbf{99.16} & \textbf{99.17} \\
\hline
\multirow{4}{*}{SqueezeNet} & $w$-based       & 62.85 & 62.17 & 60.88 \\
                            & HAQ             & 63.27 & 62.66 & 60.56 \\
                            & BO              & 62.91 & 62.24 & 60.42 \\
                            & \textbf{Ours}   & \textbf{63.47} & \textbf{62.79} & \textbf{60.90}\\
\hline
\multirow{4}{*}{TinyLLaMa}  & $w$-based       & 54.04 & 48.03 & 42.12 \\
                            & HAQ             & 55.60 & 52.77 & 44.24 \\
                            & BO              & 52.89 & 50.59 & 44.80 \\
                            & \textbf{Ours}   & \textbf{57.20} & \textbf{54.50} & \textbf{44.93} \\
\hline
\end{tabular}
\end{table}

As shown in the table, our method achieves better accuracy in most of the results across most baseline models and various BOPS constraints. For larger models with more layers and larger search spaces, e.g., SqueezeNet and TinyLLaMa, the effectiveness of other algorithms is degraded due to the vast search spaces. In contrast, our approach is able to maintain the model accuracy, thanks to our near-constraint optimization strategy, which effectively reduces search spaces and improves sample efficiency.

\begin{figure}[ht]
    \centering
    \includegraphics[width=0.48\textwidth]{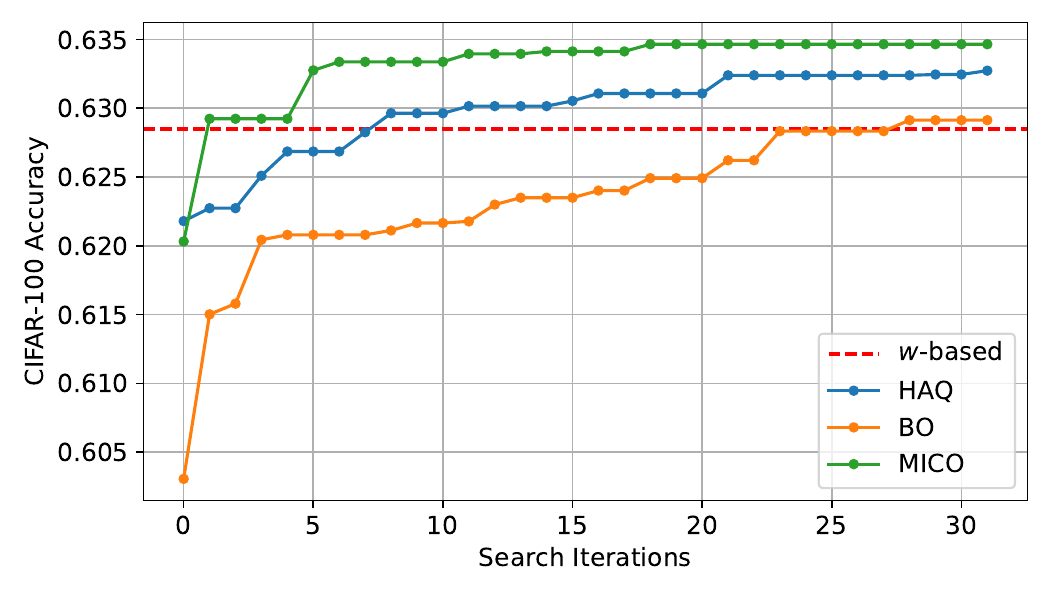}
    \caption{Comparison of PTQ Search Traces}
    \label{fig:ptq_trace}
\end{figure}

For a better demonstration of the exploration efficiency, we plot the optimization traces of algorithms in Fig.~\ref{fig:ptq_trace}, which illustrates the best accuracy achieved by each method during the PTQ search for SqueezeNet under a 0.6 BOPs constraint.

As shown in the figure, our proposed algorithm consistently improves accuracy and outperforms all other methods throughout the iterations, demonstrating the high efficiency of our method in exploring MPQ schemes with better results. Notably, our algorithm surpasses the NLP-based method at its second shot, which means that only 18 schemes are evaluated for our method, while the NLP-based method requires 26 schemes to get the layer-wise sensitivity of SqueezeNet.

We also compare the long-term search results of our method with those of the HAQ method. As shown in Fig.~\ref{fig:ptq_trace}, our method achieves an average accuracy of 63.47\% after 32 iterations, compared to HAQ's lower accuracy of 63.27\%. In the later iterations, we find that the HAQ method takes another 59 samples to finally achieve the same accuracy of 63.47\%. Since for large models like SqueezeNet, the evaluation time of networks dominates the overall search time, it means HAQ requires 1.84$\times$ more time to achieve the same result as our method.

\subsection{Mixed Precision QAT Search Results}

While many existing MPQ works focus on bit widths above 4 bits (such as 5, 6, 7 bits), the parallelism can not be effectively enhanced, since most processors and accelerators are designed with arithmetic units that operate with power-of-2 bitwidths (32-bit or 64-bit). For example, if we want to deploy a model with 5-bit weights on most of the hardware in Tab.~\ref{tab:MP_Hardware}, such as XpulpNN~\cite{xpulpnn}, the 5-bit weights need to be extended back to 8-bit before the computation, which hinders the acceleration.

Therefore, to achieve practical speedup in mixed-precision computing, we extend the design space to include bitwidths below 4 bits, which heavily rely on QAT to achieve acceptable accuracy as discussed in Sec.~\ref{sec:ptq_vs_qat}. We conduct QAT search experiments with the bitwidth choices of 1/2/4/8 bits, where each scheme is shortly re-trained during the exploration, and the final best model will be re-trained for an extended period~($\approx$ pre-trained time/10).

The search budgets for learning-based methods are set to 32 samples, consisting of 16 initial samples and 16 search samples. The QAT search results are listed in the Tab.~\ref{tab:qat search}.

\begin{table}[ht]
\centering
\begin{threeparttable}
\caption{QAT Search (1/2/4/8-bit) Accuracy Results}
\label{tab:qat search}
\begin{tabular}{ l | l | c | c | c }
\hline
\textbf{Model}              & \textbf{Method} & \textbf{0.6BOPs} & \textbf{0.5BOPs} & \textbf{0.4BOPs} \\
\hline
\multirow{4}{*}{CNN4}       & $w$-based       & 70.37 & 58.59 & 69.51 \\
                            & HAQ             & \textbf{79.31} & 78.26 & 77.66 \\
                            & BO              & 77.86 & 77.09 & 77.14  \\
                            & \textbf{Ours}   & 77.35 & \textbf{78.48} & \textbf{78.01}\\
\hline
\multirow{4}{*}{LeNet5}     & $w$-based       & 99.24 & 99.00 & \textbf{99.36} \\
                            & HAQ             & 99.29 & 99.28 & 99.26 \\
                            & BO              & 99.03 & 98.99 & 99.04 \\
                            & \textbf{Ours}   & \textbf{99.37} & \textbf{99.29} & \textbf{99.36}\\
\hline
\multirow{4}{*}{SqueezeNet} & $w$-based       & N.A.\tnote{*} & N.A.\tnote{*} & 60.18 \\
                            & HAQ  & \textbf{65.49} & 65.05 & \textbf{63.38} \\
                            & BO  & 53.57 & 53.27 & 50.98 \\
                            & \textbf{Ours}   & \textbf{65.49} & \textbf{65.35} & 62.27 \\
\hline
\multirow{4}{*}{TinyLLaMa}  & $w$-based       & 46.93          & 46.70 & 44.49 \\
                            & HAQ             & 47.28          & 49.48 & \textbf{47.41} \\
                            & BO              & 29.83          & 29.73 & 29.54 \\
                            & \textbf{Ours}   & \textbf{53.51} & \textbf{50.59} & 42.62 \\
\hline
\end{tabular}
\begin{tablenotes}
\item [*] The NLP solver fails to find a solution.
\end{tablenotes}
\end{threeparttable}
\end{table}


Shifting from a higher bitwidth search space to a lower one typically results in lower accuracies of quantized models. But thanks to QAT re-training, the accuracies are recovered and even improved for the small models compared to their 8-bit PTQ accuracy.

The $w$-based method's limitations become more apparent in the QAT setting, as it overly quantizes some layers directly to 1-bit, leading to low QAT accuracies. For SqueezeNet, which has many layers, the $w$-based method even fails to find a solution. Compared to the PTQ search results, Bayesian optimization performs much worse in the QAT scenario. Due to the high dimensionality of large models, the BO method struggles to learn from the huge MPQ spaces. HAQ achieves competitive results on large models thanks to its RL agents, which are good at making MPQ decisions layer by layer, but its performance is limited on smaller models.

Our method maintains strong performance under the QAT environment. Although the results are slightly degraded due to the complexity of large models and the QAT process, our method's overall performance is still highly competitive.

\subsection{Ablation Study}

To validate the effectiveness of our proposed techniques, we conduct ablation experiments on the PTQ search problem for SqueezeNet and TinyLLaMa under a 0.6 BOPs constraint. Starting with the basic random forest models, we incorporate the two main sampling techniques discussed in Sec.~\ref{sec:initial} and Sec.~\ref{sec:optimization} one by one to assess their impact.

\begin{table}[h]
    \centering
    \caption{Ablation Study on PTQ Search}
    \begin{tabular}{l|l|l}
    \hline
    \textbf{Method}      & \textbf{SqueezeNet Acc.} (\%) & \textbf{TinyLLaMa Acc.} (\%)\\
    \hline
    Random Forest        &  62.91                        & 53.19 \\
    + Orthogonal Initial &  62.96 (+0.05)                & 53.51 (+0.32) \\
    + Near-Constraint    &  \textbf{63.47} (+0.56)        & \textbf{57.20} (+4.01)\\
    \hline
    \end{tabular}
    \label{tab:ablation}
\end{table}

As shown in Tab.~\ref{tab:ablation}, the optimization results are first improved by the orthogonal initial samples, which enhances the initial learning of the accuracy predictors. Then, by focusing on the near-constraint spaces, the optimization results are further promoted.

\section{End-to-End MiCo Exploration Results}

To demonstrate the end-to-end exploration and deployment flow of our MiCo framework, we select two hardware platforms as the showcases, and run the complete flows on them for MPQ models. The two platforms are the aforementioned mixed-precision accelerator and the extended CPU in Fig.~\ref{fig:corr_bops}.
We run the flows with both the hardware-agnostic BOPs constraint and our hardware-aware CBOPs model constraint to show the impact of proxy models on the final hardware latencies.

\subsection{End-to-End Exploration on BitFusion Accelerator}

The BitFusion~\cite{bit_fusion} accelerator is a bit-flexible systolic array accelerator, which supports mixed-precision computations from 2-bit to 8-bit integers. For the first showcase of a complete MiCo flow, we utilize the simulator of BitFusion to set up an accelerator with 32$\times$16 processing elements as the hardware target of our end-to-end exploration. We use the linear CBOPs models for the latency prediction, as defined by Equation~\ref{eq:cbops}.

With the MiCo framework, we run the MPQ exploration on a 7-layer VGG model~\cite{vgg} trained on the CIFAR-10 dataset, and deploy MPQ models on the cycle-accurate simulator of BitFusion to evaluate the actual hardware latencies.

\begin{table}[ht]
\centering
\caption{Exploration \& Deployment for VGG on BitFusion}

\label{tab:vgg_on_bf}
\begin{tabular}{l | lll}
\hline
\textbf{Precision} &\textbf{Acc.} (\%) & \textbf{Constraint} & \textbf{Cycles} (Ratio) \\
\hline
8-bit  & 79.10 & -        & 4.15M (1.0$\times$)  \\
Mixed     & 79.09 & 0.8$\times$BOPs  & 3.62M (\textbf{\textcolor{BrickRed}{0.87$\times$}}) \\
Mixed     & 79.09 & 0.8$\times$CBOPs & \textbf{3.20M} (\textbf{\textcolor{ForestGreen}{0.77$\times$}}) \\
\hline
\end{tabular}
\end{table}

As shown in Tab.~\ref{tab:vgg_on_bf}, our framework successfully found VGG models with MPQ schemes that lead to minimal accuracy drops, under an 80\% constraint of BOPs and CBOPs proxies. The simulation results indicate that using BOPs as a proxy leads to a reduction in actual hardware latency to only 87\%. In contrast, utilizing the CBOPs model achieves a further latency reduction to 77\%. This proves the superiority of our CBOPs model over the traditional BOPs model, which is built specifically for the hardware target. The hardware awareness brought by the CBOPs lead to better satisfaction of the latency constraints and higher speedup.

\subsection{End-to-End Exploration on SIMD-Extended RISC-V CPUs}
\label{sec:e2e_on_rv}
In this experiment, we utilize SIMD-extended RISC-V CPUs as the hardware targets to demonstrate our exploration-deployment framework.

We extend the 32-bit VexiiRiscv~\cite{vexiiriscv} CPUs with 10 custom instructions (named as \textit{VexiiMiCo}), as listed in Tab.~\ref{tab:mico_isa}, to support 10 combinations between two 8/4/2/1-bit operands. All of these instructions are single-cycle R-type RISC-V instructions, which process two packed vectors from register \textit{rS1} and register \textit{rS2}, and write the result back to register \textit{rD}. With a word length of 32 bits in 32-bit CPUs, each register can contain a packed vector of low bitwidth elements, i.e., 4$\times$INT8, 8$\times$INT4, 16$\times$INT2, and 32$\times$INT1. 

\begin{table}[h]
    \centering
    \caption{VexiiMiCo - Mixed Precision SIMD Dot Product ISA}
    \begin{tabular}{|l|l|l|l|}
    \hline
    \textbf{Instruction} & \textbf{rS1 Data} & \textbf{rS2 Data} & \textbf{rD Data}\\
    \hline
    DotP.8x\{8,4,2,1\}  & 4$\times$8-bit  & 4$\times$\{8,4,2,1\}-bit & \multirow{4}{*}{32-bit Res.} \\
    DotP.4x\{4,2,1\}    & 8$\times$4-bit  & 8$\times$\{4,2,1\}-bit &\\
    DotP.2x\{2,1\}      & 16$\times$2-bit & 16$\times$\{2,1\}-bit &\\
    DotP.1x1            & 32$\times$1-bit & 32$\times$1-bit &\\
    \hline
    \textbf{Computation} &\multicolumn{3}{c|}{rD=rS1[0]*rS2[0]+$\cdots$+rS1[N]*rS2[N]} \\
    \hline
    \end{tabular}
    \label{tab:mico_isa}
\end{table}

Similar to the CPU extensions listed in Tab.~\ref{tab:MP_Hardware}, such as MPIC~\cite{MPIC}, our custom instructions perform standard dot-product operations between two vectors, which are heavily involved in many neural network operations, especially matrix multiplication (MatMul). Particularly, we use \textit{rS1} to store elements with higher or the same bitwidths as those contained in \textit{rS2}. When the elements in \textit{rS2} have lower bit widths, they will be extracted and extended to match the bit widths of \textit{rS1}.

We benchmark the VexiiMiCo on a $64\times64\times64$ MatMul computation on different precisions. Compared to the basic VexiiRiscv CPU with RV32IMFC ISA support, our VexiiMiCo CPU with SIMD extension achieves a $3.7\times$ faster INT8 MatMul computation and can increase the acceleration up to $19.2\times$ with reduced bitwidths.

To exhibit MiCo's flexibility in adapting to different architectures, we choose 3 configurations of the base CPUs with different pipeline widths, cache sizes, and branch predictors, which will lead to varied overall performances. We denote 3 CPU configurations as \textit{Tiny}, \textit{Small} and \textit{High}:
\begin{itemize}
    \item \textit{Tiny}: a single-issue CPU without any cache or branch predictor.
    \item \textit{Small}: a single-issue CPU with I-cache, D-cache, and Branch Target Buffer~(BTB).
    \item \textit{High}: a dual-issue CPU with larger I-cache, D-cache, and a gshare branch predictor.
\end{itemize}
All 3 CPUs are extended with the hardware unit and the ISA in Tab.~\ref{tab:mico_isa} to accelerate the MPQ inference. We perform MPQ exploration on the CNN4 and LeNet5 in Tab.~\ref{tab:networks}, deploy MPQ models on the above CPUs, and evaluate the actual latencies via RTL simulations. As explained in previous sections, we adopt QAT flow to recover the accuracy for the bitwidth space that includes 1-/2-bit choices.

\begin{table}[ht]
\centering
\caption{Exploration \& Deployment for CNNs on VexiiMiCo CPUs}
\label{tab:CNN4 on CPU}
\begin{tabular}{l|l|lll}
\hline
\textbf{Type} & \textbf{Precision} &\textbf{Acc.} (\%) & \textbf{Constraint} & \textbf{Cycles} (Ratio) \\
\hline
\multirow{3}{*}{\textit{Tiny}} & 8-bit  & 77.87 & -        &  67.4M (1.0$\times$)  \\
& Mixed (QAT) & 77.42  & 0.8$\times$BOPs   &  62.8M (0.93$\times$) \\
& Mixed (QAT) & 76.43  & 0.8$\times$CBOPs  &  \textbf{54.7M (0.81$\times$)}\\
\hline
\multirow{3}{*}{\textit{Small}} & 8-bit  & 77.87 & -        & 69.4M (1.0$\times$)  \\
& Mixed (QAT) & 77.42  & 0.8$\times$BOPs   & 66.2M (0.95$\times$) \\
& Mixed (QAT) & 76.57  & 0.8$\times$CBOPs  & \textbf{57.1M (0.82$\times$)} \\
\hline
\multirow{3}{*}{\textit{High}} & 8-bit (QAT)  & 77.87 & -        & 48.1M (1.0$\times$)  \\
& Mixed (QAT) & 77.42 & 0.8$\times$BOPs  & 46.6M (0.97$\times$) \\
& Mixed (QAT) & 76.88 & 0.8$\times$CBOPs & \textbf{40.7M (0.85$\times$)} \\
\hline
\end{tabular}
\end{table}

\begin{table}[ht]
\centering
\caption{Exploration \& Deployment for LeNet5 on VexiiMiCo CPUs}
\label{tab:LeNet5 on CPU}
\begin{tabular}{l|l|lll}
\hline
\textbf{Type} & \textbf{Precision} &\textbf{Acc.} (\%) & \textbf{Constraint} & \textbf{Cycles} (Ratio) \\
\hline
\multirow{3}{*}{\textit{Tiny}} & 8-bit   &  99.37 & -        &  5.12M (1.0$\times$)  \\
& Mixed (QAT) & 98.68 & 0.8$\times$BOPs  &  4.82M (0.94$\times$) \\
& Mixed (QAT) & 96.57 & 0.8$\times$CBOPs &  \textbf{4.20M (0.82$\times$)}\\
\hline
\multirow{3}{*}{\textit{Small}} & 8-bit  &  99.37 & -        & 4.80M (1.0$\times$)  \\
& Mixed (QAT) & 98.68 & 0.8$\times$BOPs  & 4.73M (0.98$\times$) \\
& Mixed (QAT) & 98.28 & 0.8$\times$CBOPs & \textbf{3.95M (0.82$\times$)} \\
\hline
\multirow{3}{*}{\textit{High}} & 8-bit &  99.37 & - & 3.31M (1.0$\times$)  \\
& Mixed (QAT) & 98.68 & 0.8$\times$BOPs         & 3.08M (0.93$\times$) \\
& Mixed (QAT) & 97.04 & 0.8$\times$CBOPs        & \textbf{2.64M (0.80$\times$)} \\
\hline
\end{tabular}
\end{table}

Hardware latency on CPUs can exhibit greater non-linearity compared to accelerators due to factors like caching and operations such as im2col or dynamic quantization. Therefore, as explained in Sec.~\ref{sec:cbops_proxy}, a mixture of linear models and regression trees is used for the CBOPs model of extended CPUs.

As shown in Tab.~\ref{tab:CNN4 on CPU} and Tab.~\ref{tab:LeNet5 on CPU}, although optimizing with BOPs constraints can lead to speed up as well, the improvement is much lower than expected. With hardware-dependent CBOPs models for each CPU configuration, the constraints successfully guide the exploration towards higher speedup, leading to lower end-to-end latency for the MPQ model deployment.

\section{Conclusion}

We proposed MiCo, an end-to-end framework featuring an efficient, hardware-aware MPQ search algorithm and a direct deployment flow. Utilizing ensemble model-based optimization, novel sampling techniques, and a hardware-aware CBOPs latency proxy, MiCo effectively explores the MPQ space to find high-accuracy models that meet constraints, outperforming conventional solutions. Our end-to-end deployment path, facilitated by Python APIs, converts these optimized models into practical speedups in real edge AI scenarios on diverse hardware like accelerators and extended RISC-V CPUs.

\bibliography{refs}
\bibliographystyle{ieeetr}
\vspace{12pt}
\end{document}